\documentclass[letterpaper]{article} 
\usepackage{aaai2026}  
\usepackage{times}  
\usepackage{helvet}  
\usepackage{courier}  
\usepackage[hyphens]{url}  
\usepackage{graphicx} 
\usepackage{natbib}  
\usepackage{caption} 
\usepackage{algorithm}
\usepackage{algorithmic}
\usepackage{amsmath, amsthm, amssymb}
\usepackage{multirow}
\usepackage{makecell}

\nocopyright
\usepackage{bbding}
\usepackage{newfloat}
\usepackage{listings}
\newcommand \blfootnote[1]{
    \begingroup
        \renewcommand
        \thefootnote{}\footnote{#1}
        \addtocounter{footnote}{-1}
        \vspace{-1ex}
    \endgroup
}

\DeclareCaptionStyle{ruled}{labelfont=normalfont,labelsep=colon,strut=off} 
\floatstyle{ruled}
\newfloat{listing}{tb}{lst}{}
\floatname{listing}{Listing}
\title{MMG-Vid: Maximizing Marginal Gains at Segment-level \\ and Token-level for Efficient Video LLMs 
}
\author{Junpeng Ma$^{1,2}$, Qizhe Zhang$^{1}$, Ming Lu$^{1}$$^{\dagger}$, Zhibin Wang$^{3}$$^{\dagger}$, \\ Qiang Zhou$^{3}$, Jun Song$^{3}$, Shanghang Zhang$^{1}$$^{\text{\Envelope}}$
}
\affiliations {
{$^1$ State Key Laboratory of Multimedia Information Processing, School of Computer Science, Peking University},\\
{$^2$ Fudan University},
{$^3$ Taobao \& Tmall Group of Alibaba}
}

\usepackage{bibentry}

\begin{document}
\maketitle

\blfootnote{$\dagger$ Project leader.\quad $^{\text{\Envelope}}$ Corresponding author. }

\begin{abstract}
Video Large Language Models (VLLMs) excel in video understanding, but their excessive visual tokens pose a significant computational challenge for real-world applications. Current methods aim to enhance inference efficiency by visual token pruning. However, they do not consider the dynamic characteristics and temporal dependencies of video frames, as they perceive video understanding as a multi-frame task. To address these challenges, we propose \textbf{MMG-Vid}, a novel training-free visual token pruning framework that removes redundancy by \textbf{M}aximizing \textbf{M}arginal \textbf{G}ains at both segment-level and token-level. Specifically, we first divide the video into segments based on frame similarity, and then dynamically allocate the token budget for each segment to maximize the marginal gain of each segment. Subsequently, we propose a temporal-guided DPC algorithm that jointly models inter-frame uniqueness and intra-frame diversity, thereby maximizing the marginal gain of each token. By combining both stages, MMG-Vid can maximize the utilization of the limited token budget, significantly improving efficiency while maintaining strong performance. Extensive experiments demonstrate that MMG-Vid can maintain over \textbf{99.5\%} of the original performance, while effectively \textbf{reducing 75\%} visual tokens and accelerating the prefilling stage by \textbf{3.9x} on LLaVA-OneVision-7B. Code is available at \url{https://github.com/ylnxxts/MMG-Vid}.

\end{abstract}

\section{Introduction}
\label{sec:introduction}
Recently, Video Large Language Models (VLLMs) have shown impressive abilities in understanding video content~\cite{li2024llava-ov,zhang2024llava-video}, which typically rely on densely sampling video frames, resulting in a substantial volume of visual tokens~\cite{zhang2024longva, chen2024longvila}. However, processing these visual tokens is computationally demanding due to the quadratic complexity of the self-attention mechanism~\cite{vaswani2017transformer,liu2025shifting}. This bottleneck leads to substantial inference latency and memory overhead, severely hindering their deployment in resource-constrained scenarios. Consequently, developing token compression algorithms to effectively reduce redundancy while preserving critical semantic information has become a primary focus of current research~\cite{tao2025dycoke,huang2024prunevid,fu2024framefusion}.

\begin{figure*}[t]
    \centering
    \includegraphics[width=\textwidth]{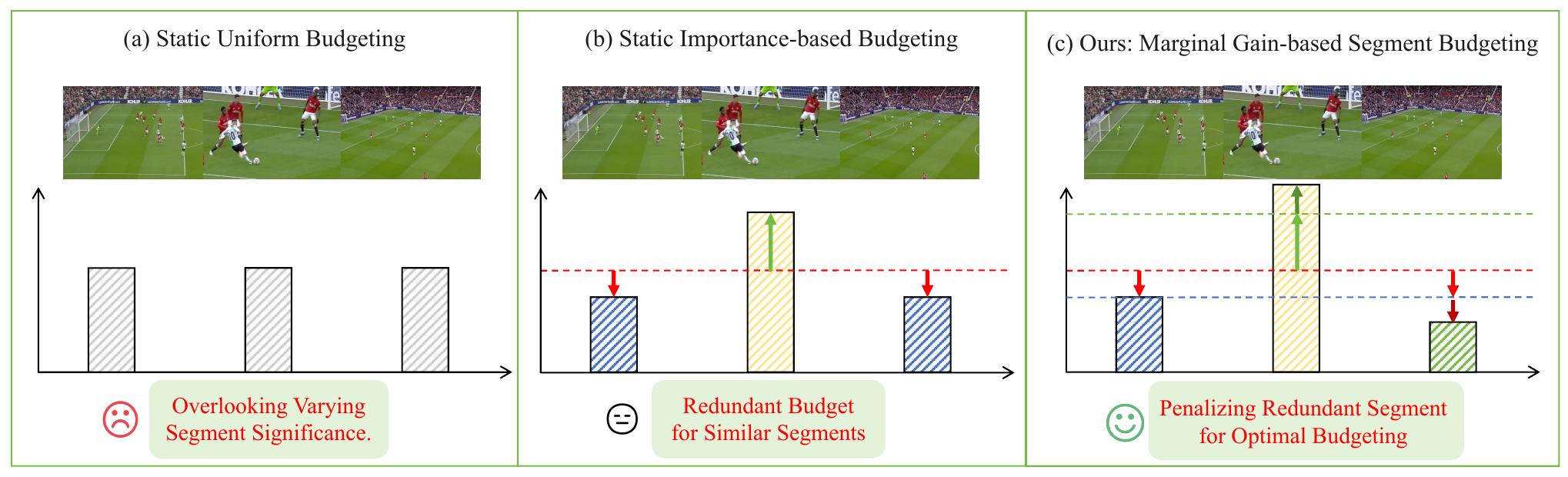}
    \caption{\textbf{Comparison of different budget allocation strategies.} (a) Static uniform budgeting overlooks varying segment significance. (b) Static importance-based budgeting acknowledges the significance of segments but wastes resources by allocating redundant budgets to two visually similar static segments (the first and third). (c) Our marginal gain-based segment budgeting further reduces redundancy by penalizing a segment's budget if its information is already included in previously selected segments. This results in an optimal allocation of the budget that takes into account the dynamic characteristics.}
\label{fig:introduction}
\end{figure*}

Current methods do not effectively consider the dynamic characteristics and temporal dependencies of video frames, as they view video understanding as a multi-frame task, leading to suboptimal performance. The shortcomings of previous works can be summarized from two perspectives: 1) Static token budget: As shown in Figure~\ref{fig:introduction}, previous approaches typically assign a static token budget to all video segments, disregarding the dynamic characteristics~\cite{tao2025dycoke,shen2025fastvid,sun2025llava-scissor}. For example, a complex segment with multiple human interactions carries significantly richer semantic information than a relatively static landscape shot, and therefore should be allocated more computational resources. While recent works like VidCom$^2$~\cite{liu2025vidcom2} attempt to allocate token budgets based on the importance, their static allocation strategy fails to reflect the inherently dynamic characteristics. The importance of a segment should not be considered static but rather dynamically adjusted based on information already retained. We argue that the true value of a segment lies in \textbf{maximizing segment-level marginal gain}. 2) Disjoint pruning strategy: Most prior works treat diversity and importance as two separate metrics~\cite{huang2024prunevid,fu2024framefusion,shao2025holitom}. A typical pruning pipeline involves two stages: temporal redundancy removal and spatial importance selection, executed in a strictly sequential manner. However, a token that is spatially important may be discarded too early in the first stage, while a token kept in the second stage may not be globally optimal. We contend that token selection should also consider the temporal dependencies of video frames, aiming for globally optimal pruning by \textbf{maximizing token-level marginal gain}.

To address these limitations in a unified manner, we propose MMG-Vid, a novel training-free video token pruning framework that \textbf{M}aximizes \textbf{M}arginal \textbf{G}ains at segment-level and token-level to preserve critical information while reducing computational cost. MMG-Vid consists of three stages:
\textbf{(i) Similarity-based Frame Segmentation:}
Long videos often contain multiple discontinuous segments, and simply pruning individual frames does not take advantage of the temporal structure inherent in videos. To address this, we first divide the video into semantically coherent segments by analyzing the similarities between adjacent frames. This approach allows us to prune more effectively by leveraging the temporal relationships within each segment.
\textbf{(ii) Marginal Gain-based Segment Budgeting:}
As previously mentioned, segments differ in their dynamic characteristics, and distributing the budget evenly can lead to the loss of important information in high-density segments. To tackle this issue, we propose an iterative dynamic budget allocation process. At each step, it evaluates the marginal gain of the current segment, ensuring that we make the best use of limited computational resources at the segment level.
\textbf{(iii) Marginal Gain-based Token Pruning:}
To address the limitations of disjoint token pruning, we propose a new approach called temporal-guided density peak clustering (TG-DPC). This algorithm models both inter-frame distinctiveness and intra-frame diversity, allowing it to effectively reduce spatiotemporal redundancy in a progressive manner. Ultimately, TG-DPC aims to maximize marginal gains at the token level.

We apply our MMG-Vid to the most widely used VLLMs, LLaVA-Video and LLaVA-OneVision, and evaluate on multiple video question answering benchmarks. 
Empirical results demonstrate that MMG-Vid substantially reduces the computational overhead of VLLMs while robustly preserving their performance across a wide range of retention ratios. For instance, when applied to LLaVA-OneVision-7B, MMG-Vid prunes \textbf{75\%} of video tokens, achieving a \textbf{3.9x} speedup in the prefilling stage, all while maintaining \textbf{99.5\%} of the model's original performance on average. 

The key contributions are summarized as follows:
\begin{itemize}
     \item We reformulate video token pruning as a problem of maximizing marginal gain, addressing the limitations of previous methods at both segment-level and token-level.
     \item We present MMG-Vid, a novel framework for training-free video token pruning that utilizes marginal gain-based segment budgeting and token pruning to efficiently manage limited computational resources.
     \item Our MMG-Vid achieves state-of-the-art performance on both LLaVA-Video and LLaVA-OneVision accross various benchmarks, substantially improving inference efficiency while largely maintaining original performance.
\end{itemize}

\section{Related Work}
\label{sec:related_work}

\subsection{Video Large Language Models}
Typical Video Large Language Models (VLLMs) employ visual encoders and projectors to independently encode each frame into visual tokens, which are then concatenated with the user query and fed into the LLM. Recent studies have explored diverse methods for advancing video understanding~\cite{song2024moviechat,zhang2025videollama}.
LongVA~\cite{zhang2024longva} enhances long-video understanding by extending the LLM's context length. Qwen2-VL~\cite{wang2024qwen2-vl} improves temporal awareness using M-RoPE. LLaVA-OneVision~\cite{li2024llava-ov} unifies image and video tasks and efficiently compresses tokens by bilinear interpolation. LLaVA-Video~\cite{zhang2024llava-video}  uses newline tokens for spatial-temporal grounding. However, excessive video frames yield a vast quantity of visual tokens, limiting the practical application of VLLMs due to the quadratic computational complexity~\cite{vaswani2017transformer}.

\subsection{Token Compression for VLLMs}

Token compression directly prunes visual tokens to improve inference efficiency in VLLMs. Although training-aware paradigms~\cite{li2024llama-vid,ye2025voco-llama,zhang2025llava-mini,yang2025visionthink} effectively reduce sequence length while maintaining model performance, they require modifications during training and are thus computationally expensive. Consequently, an increasing number of works have begun exploring training-free approaches for enhancing inference efficiency~\cite{zhang2024sparsevlm,xing2024pyramiddrop,shang2024llava-prumerge,han2024ficoco,liu2025globalcom2,zhang2024vispruner,zhang2025cdpruner}. FastV~\cite{chen2024fastv} first identifies inefficient cross-modal attention within the language model and evaluates the importance of visual tokens based on the attention received from text tokens. VisionZip~\cite{yang2025visionzip} selects important tokens using attention from the [CLS] token in the visual encoder and then merges the remaining ones. DivPrune~\cite{alvar2025divprune} formulates the token pruning problem as a Max-Min Diversity Problem and iteratively retains visual tokens using a greedy algorithm.
However, none of the aforementioned methods are designed specifically for video understanding and ignore the temporal connections between different video frames. To address this, PruneVid~\cite{huang2024prunevid} clusters frames into segments and classifies tokens within each segment as either static or dynamic, applying different pruning strategies accordingly. FrameFusion~\cite{fu2024framefusion} first merges tokens across frames and then selects important tokens. 
Inspired by DivPrune, we propose a training-free video pruning strategy that maximizes the marginal gains at both segment-level and token-level, enabling more effective use of limited computational resources while maintaining overall performance.

\begin{figure*}[t]
    \centering
    \includegraphics[width=\textwidth]{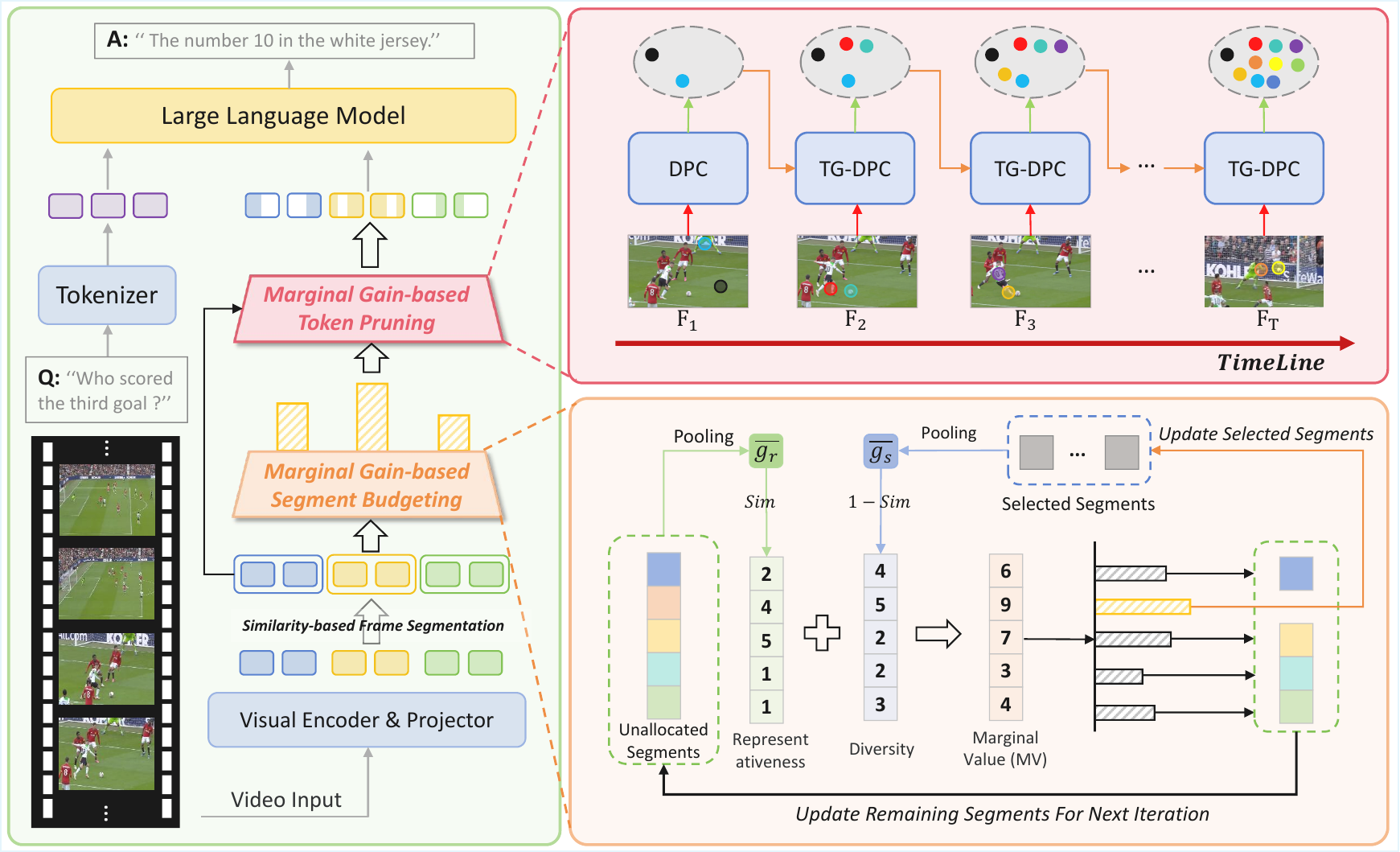}
    \caption{\textbf{Overall framework.} \textbf{Segment-level (Bottom-Right): } We iteratively calculate the marginal gain (a combination of representativeness and diversity) for each segment to dynamically allocate budget, prioritizing more informative segments. \textbf{Token-level (Top-Right): } Our proposed TG-DPC progressively prunes each frame by selecting tokens that are both salient within the frame and novel across the temporal dimension, guided by the set of previously selected tokens } 
    \vspace{-4mm}
    \label{fig:method}
\end{figure*}

\section{Methodology}
In this section, we formulate the token pruning task as a constrained optimal subset selection problem, and present our MMG-Vid framework to derive a feasible solution. MMG-Vid employs the principles of Maximal Marginal Gains, primarily comprising three components: 1) Similarity-based Frame Segmentation, 2) Marginal Gain-based Segment Budgeting, and 3) Marginal Gain-based Token Pruning.
\subsection{Problem Formulation}

First, we claim that the essence of visual token pruning is to identify and select an optimal subset $\mathcal{S}_{\text{sub}}$ from a given visual token set $\mathcal{S}_{\text{v}}$ that maximizes a predefined quality function, under the constraint of a specific retention ratio $R$. We formally model this as a constrained subset selection problem. 

Let $V$ denote a video, represented by a set of tokens $\mathcal{T}_v = \{t_1, t_2, \dots, t_N\}$, where $N$ is the total number of tokens across all frames.  We first define a quality function for a token subset that jointly quantifies the representativeness and diversity of each token within it, as expressed in Eq~\eqref{eq:1}:

\begin{equation} 
\label{eq:1}
Q(\mathcal{T}_{\text{sub}}) = 
\underbrace{\sum_{t_i \in \mathcal{T}_{\text{sub}}} I(t_i)}_{\text{Total Representativeness}} - \beta 
\underbrace{\sum_{\substack{t_i, t_j \in \mathcal{T}_{\text{sub}} \\ i \neq j}} D(t_i, t_j)}_{\text{Total Redundancy}}
\end{equation}

Our optimization objective is to find an optimal subset $\mathcal{T}^{*}_{\text{sub}}$ that maximizes the quality function, subject to a constraint determined by the pre-defined retention ratio $R$: 

\begin{equation}
\begin{aligned}
\mathcal{T}^{*}_{\text{sub}} = \underset{\mathcal{T}_{\text{sub}} \subseteq \mathcal{T}_{\text{v}}}{\text{argmax}} & \quad Q(\mathcal{T}_{\text{sub}}) \\
\text{subject to} \quad & |\mathcal{T}_{\text{sub}}| \leq R\cdot N
\end{aligned}
\end{equation}

This is a variant of the Max-Min Diversity Problem, which is known to be NP-hard~\cite{porumbel2011mmdp,parreno2021measuringmdp}. A brute-force search over all possible subsets is computationally intractable. Given the infeasibility of direct global optimization, we resort to a three-stage framework under the guidance of the MMG principle to find an approximate solution.

\subsection{MMG-Vid} 

 We propose a novel three-stage framework, MMG-Vid, an efficient approximation approach by decomposing the complex global token selection problem into a series of more tractable sub-problems, which are in turn solved via our proposed three-stage approach.

\subsection{Stage-1: Similarity-based Frame Segmentation}
We begin by reformulating the problem space. Motivated by the observation that video content is non-uniformly distributed across the temporal dimension, processing the entire video as a monolithic, unstructured token set fails to exploit its inherent semantic structure. To align our approach with this intrinsic structure, we first segment the video into $K$ semantically coherent segments based on temporal dynamics.

Let $f_t \in \mathbb{R}^d$ be the feature embedding of frame $F_t$ obtained by average pooling its corresponding visual tokens. We compute the cosine similarity between adjacent frames:

\begin{equation} 
\begin{aligned}
\label{eq:3}
\text{sim}(F_t, F_{t+1}) = \frac{f_t \cdot f_{t+1}}{\|f_t\| \|f_{t+1}\|} \\
B=\{i|\text{sim}(F_t, F_{t+1})<\tau\}
\end{aligned}
\end{equation}

We define an initial set of segment boundaries $B$ by identifying frames where the similarity to the next frame drops below a threshold $\tau$. An initial set of segments $\{S_1, S_2, ...\}$ is thus formed. To ensure temporal coherence and avoid fragmented, single-frame segments, we enforce a minimum segment length. Any segment $S_k$ with $|S_k| = 1$ is merged into its most similar adjacent segment:
\begin{equation} \label{eq:eq4}
\begin{aligned}
    S_{k^*} &\leftarrow S_{k^*} \cup S_k, \\
    \text{where } k^* &= \underset{j \in \{k-1, k+1\}}{\operatorname{arg\,max}} \, \operatorname{sim}(S_k, S_j)
\end{aligned}
\end{equation}
This process yields a final set of $K$ segments, $\{S_1, S_2, ..., S_K\}$, where each segment represents a temporally and visually consistent segment of the video.

\subsection{Stage-2: Marginal Gain-based Segment Budgeting}

After segmenting the video into clips in Stage-1, the problem is thus reformulated as selecting optimal token subsets from each segment, thereby decomposing the overall task into a set of independent sub-problems. Prior works typically assign a uniform token budget to each segment and then perform intra-segment pruning, overlooking the inherent heterogeneity and inter-dependencies among them. Instead, to avoid losing critical information in content-dense segments while wasting the computational budget on content-sparse ones, we propose a budget allocation strategy based on dynamic marginal value, aiming to devise an optimal token allocation strategy within a constrained total budget, such that the resulting subset of tokens can maximize the semantic diversity and coverage of the entire video. 

First, to prevent critical information loss, we establish a universal minimum retention ratio $R_{min}$ for all segments. This guarantees a baseline representation for every segment, and the remaining discretionary token budget is calculated as $R_{extra} = R - R_{min}$, which is allocated iteratively based on the marginal gain of each segment.

Then, we argue that a segment's value is not absolute but is relative to the information already processed. Therefore, we sequentially and iteratively determine an optimal ordering and importance score for each segment. We begin with an empty set of selected segments. At each step, for all remaining unselected segments, we compute a Marginal Value (MV) to represent the optimal trade-off between the segment's intrinsic relevance and its diversity with respect to the set of segments already selected.

Let $\mathbf{g}_k$ be the representation of segment $S_k$, obtained by averaging all tokens within the segment. Additionally, let $\mathbb{S}_{\text{sel}}$ be the set of segments for which the budget has been allocated, and $\mathbb{S}_{\text{rem}}$ be the set of segments that remain unselected. We posit that the Marginal Value (MV) of a segment is co-determined by its representativeness, defined as its thematic relevance to the remaining segments, and its diversity, which reflects its capacity to introduce novel information:

\begin{equation} \label{eq:mv_aligned}
\begin{aligned}
&    \mathrm{MV}(S_k | \mathbb{S}_{\text{sel}}) = \lambda  \underbrace{\text{sim}(\mathbf{g}_k, \overline{\mathbf{g}_{r}})}_{\text{Representativeness}} + (1-\lambda) \underbrace{(1 - \text{sim}(\mathbf{g}_k,\overline{\mathbf{g}_{s}}))}_{\text{Diversity}} \\
&\text{where} \quad \overline{\mathbf{g}_{r}} = \frac{1}{|\mathbb{S}_{\text{rem}} |} \sum_{S_i \in \mathbb{S}_{\text{rem}}} \mathbf{g}_i, \quad \overline{\mathbf{g}_{s}} = \frac{1}{|\mathbb{S}_{\text{sel}} |} \sum_{S_j \in \mathbb{S}_{\text{sel}}} \mathbf{g}_j.
\end{aligned}
\end{equation}
$\lambda$ is a parameter to balance representativeness and diversity. The segment with the highest MV is then chosen to compute its token budget via Z-score (illustrated in eq~\eqref{eq:eq6}). 
\begin{equation}
\label{eq:eq6}
    R_k = R_{min} + R_{extra} \cdot \frac{\text{MV}(S_k) - mean(\text{MV})}{std( \text{MV})}
\end{equation}

This process is repeated until all segments are allocated. Crucially, the marginal value $\text{MV}(S_k)$ utilized in the allocation is not a pre-computed, static metric but a dynamic score determined at the precise selection moment of segment $S_k$. This ensures that the extra budget $R_{extra}$ is allocated based on a segment's actual marginal contribution to the video's overall information content, thereby maximizing information density under a given computational constraint.

\subsection{Stage-3: Marginal Gain-based Token Pruning}

After determining the token budget, we proceed to the fine-grained token pruning within each segment. Simply applying a pruning algorithm to each frame independently disregards the inherent temporal coherence of videos, leading to two primary issues: \textbf{1) Information Redundancy:} Static regions, such as backgrounds, may be repeatedly selected across consecutive frames, yielding highly similar tokens. \textbf{2) Loss of Dynamic Information:} Subtle yet crucial changes, such as object movements or pose alterations, might be overlooked if are not sufficiently salient within a single frame.

To overcome these limitations, we propose a dynamic selection strategy based on temporal information gain that treats the initial frame as a visual anchor and progressively enriches the visual information in subsequent frames.

\textbf{First Frame Token Pruning:} For the first frame in the segment, we employ the standard DPC-KNN algorithm to select an optimal token subset. For each token $t_i$,  we compute its local density:
\begin{equation}
\rho_{i}=\exp \left(-\frac{1}{n} \sum_{t_{j} \in \mathrm{kNN}\left({t}_{i}\right)} 
\left\|t_i-t_j\right\|_{2}^{2},
\right),
\label{eq:rho}
\end{equation}
where $\mathrm{kNN}(t_i)$ denotes the k-nearest neighbors of token $t_i$. Then, for each token, we compute its minimum distance $\delta_i$ to any other token with higher density : 
\begin{equation}
\label{eq:8}
\delta_{i} = \left\{
\begin{array}{ll}
  \displaystyle\min_{j:\rho_j > \rho_i} & \|t_i - t_j\|_{2}^{2}, \quad \text{if } \exists j \text{ s.t. } \rho_{j} > \rho_{i}, \\[2.5ex]
  \displaystyle\max_{j} & \|t_i - t_j\|_{2}^{2}, \quad \text{otherwise,}
\end{array}
\right.
\end{equation}
where each token score is $\gamma_i = \rho_i \cdot \delta_i $. Tokens with higher scores are selected as initial cluster centers, which are both representative (high $\rho$) and unique (high $\delta$). We select $R_k \cdot M$ tokens with the highest score ($M$ is the token number of each frame) and add them to the selected set $\mathcal{T}^{guide}_k$.

\textbf{Subsequent Frame Token Pruning:} Starting from the second frame, we introduce the \textbf{Temporal-Guided-DPC (TG-DPC)} algorithm. For each candidate token in the current frame $F_i$, we redefine its density and distance properties using the set of previously selected tokens $\mathcal{T}^{guide}_k$.

\noindent• \textbf{Temporal Relevance Density ($\rho^t$):} 
We replace the conventional local density with temporal relevance density. The $\rho^t_i$ of token $i$ is no longer determined by its neighbors in the current frame but by the historically selected tokens. 
\begin{equation}
    \rho^t_i = 1 - \exp\left(-\frac{1}{k}\sum_{j=1}^{k} \left\|t_i-t_g^{(j)}\right\|_{2}^{2}\right)
\end{equation}
where $t_g^{(j)}$ is the $j$-th nearest token to $t_i$ in $\mathcal{T}^{guide}_k$. It suggests that if a token differs from the selected tokens, the marginal gain is significant and should be chosen.

\noindent• \textbf{Intra-Frame Separation ($\delta^t$):} 
We retain the concept in standard DPC-KNN, but its calculation is now based on our new temporal relevance density $\rho^t$. For a token $t_i$, it intra-frame separation $\delta^t_i$ is defined as the minimum distance to any other token $t_m$ within the same frame that has a higher $\rho^t$ (i.e., $\rho^t_m > \rho^t_i $) :

\begin{equation}
\label{eq:9}
\delta_{i}^t = \left\{
\begin{array}{ll}
  \displaystyle\min_{m:\rho_m^t > \rho_i^t} & \|t_i - t_m\|_{2}^{2}, \quad \text{if } \exists m \text{ s.t. } \rho_m^t > \rho_i^t, \\[2.5ex]
  \displaystyle\max_{m} & \|t_i - t_m\|_{2}^{2}, \quad \text{otherwise.}
\end{array}
\right.
\end{equation}
Therefore, the final score of each token is defined as $\gamma_i^t = \rho_i^t \cdot \delta_i^t $. A higher score indicates that a token is both novel (dissimilar to past information) and core (distinct within the current frame). We then also select the top $R_k \cdot M$ tokens based on the new score $\gamma_i^t$ and upadte the selected set $\mathcal{T}^{guide}_k \leftarrow \mathcal{T}^{guide}_k\bigcup \{\mathcal{T}^{*}_i\}$. This temporally-aware selection process greedily builds a diverse set of tokens across the entire segment and yields two key benefits:

\textbf{1. Capturing Dynamics:} When an object in the segment moves or deforms (e.g., a person waving their hand), the tokens with significant marginal gain will be identified and preserved due to their novelty compared to those from the previous static state.

\textbf{2. Background Completion:} If the tokens chosen from the initial frame focus on a foreground object, the algorithm will give a higher $\rho^t_i$ value to background tokens in later frames due to their significant dissimilarity, thereby ensuring their inclusion.

\begin{table*}[t]
\centering
\renewcommand{\arraystretch}{1.05}
\setlength{\tabcolsep}{1mm}
{
\begin{tabular}{c|ccccccc|cc}
\hline
\multirow{2}{*}{Method} & \multirow{2}{*}{MVBench} & \multirow{2}{*}{\makecell{LongVideo\\Bench}} & \multirow{2}{*}{MLVU} & \multicolumn{4}{c}{VideoMME} & \multicolumn{2}{|c}{\multirow{2}{*}{Average}} \\
\cline{5-8}
& & & & Overall & Short & Medium & Long & & \\
Duration & 16s & 1$\sim$60min & 3$\sim$120min & 1$\sim$60min & 1$\sim$3min & 3$\sim$30min & 30$\sim$60min & Score & \% \\ \hline
LLaVA-Video                      & 59.2& 58.9& 67.4& 64.3& 77.3& 62.3& 53.2& 62.4& 100.0\\  \hline
\multicolumn{10}{c}{\textit{\textbf{Retention Ratio: ~25\%}}} \\ \hline

FastV{\small\texttt{(ECCV24)}}              & 52.1& 54.8& 57.8& 58.6& 68.7& 58.4& 48.7& 55.8& 89.4
\\
VisionZip{\small\texttt{(CVPR2025)}}        & 56.0& \underline{58.0}& 64.4& \underline{61.9}& \underline{73.4}& \textbf{60.6}& 51.6& 60.1& 96.3
\\
PruneVid{\small\texttt{(ACL2025)}}          & 55.0& 57.9& 64.1& 60.5& 72.2& 58.6& 50.7& 59.4& 95.2
\\
FrameFusion{\small\texttt{(ICCV2025)}}  & \underline{56.5}& 57.7& \underline{65.9}& 61.3& 72.8& 59.3& \underline{51.9}& \underline{60.4}& \underline{96.8}\\
MMG-Vid                                               & \textbf{57.6}& \textbf{58.8}& \textbf{66.2}& \textbf{62.3}& \textbf{74.7}& \underline{60.0}& \textbf{52.1}& \textbf{61.2}& \textbf{98.1}\\ \hline
\multicolumn{10}{c}{\textit{\textbf{Retention Ratio: ~20\%}}} \\ \hline

FastV{\small\texttt{(ECCV24)}}              & 50.8& 52.4& 55.2& 57.3& 65.8& 57.2& 48.8& 53.9& 86.4
\\
VisionZip{\small\texttt{(CVPR2025)}}        & \underline{55.7}& 56.3& 63.3& \underline{60.9}& 71.1& 59.4& \textbf{52.2}& 59.1& 94.7
\\
PruneVid{\small\texttt{(ACL2025)}}          & 54.6& 57.3& 63.3& 60.2& \underline{73.0}& 58.2& 49.2& 58.9& 94.4
\\
FrameFusion{\small\texttt{(ICCV2025)}}  & 55.3& \underline{57.4}& \underline{63.9}& 60.8& 72.2& \underline{59.8}& 50.3& \underline{59.4}& \underline{95.2}\\
MMG-Vid                                              & \textbf{57.2}& \textbf{58.6}& \textbf{64.9}& \textbf{61.4}& \textbf{73.4}& \textbf{60.2}& \underline{50.6}& \textbf{60.5}& \textbf{97.0}\\ \hline
\multicolumn{10}{c}{\textit{\textbf{Retention Ratio: ~15\%}}} \\ \hline

FastV{\small\texttt{(ECCV24)}}                   & 46.9& 49.8& 53.4& 54.0& 60.9& 54.4& 46.7& 51.0& 81.7
\\
VisionZip{\small\texttt{(CVPR2025)}}        & \underline{55.2}& \underline{56.3}& 62.2& \underline{60.3}& 70.3& \underline{58.8}& \textbf{51.7}& \underline{58.5}& \underline{93.8}\\
PruneVid{\small\texttt{(ACL2025)}}          & 54.0& 56.2& \underline{63.3}& 59.4& \underline{71.2}& 57.7& 49.3& 58.2& 93.3
\\
FrameFusion{\small\texttt{(ICCV2025)}} & 54.3& 54.9& 61.3& 59.3& 69.9& 58.3& 49.7& 57.5& 92.1
\\
MMG-Vid                                               & \textbf{56.1}& \textbf{57.7}& \textbf{64.8}& \textbf{61.1}& \textbf{72.3}& \textbf{60.1}& \underline{50.8}& \textbf{59.9}& \textbf{96.0}\\ \hline
\multicolumn{10}{c}{\textit{\textbf{Retention Ratio: ~10\%}}} \\ \hline

FastV{\small\texttt{(ECCV24)}}                   & 43.2& 46.5& 53.1& 49.6& 54.0& 50.3& 44.6& 48.1& 77.1
\\
VisionZip{\small\texttt{(CVPR2025)}}        & \underline{53.8}& 52.9& 60.3& \underline{58.7}& 67.4& \underline{57.7}& \textbf{51.1}& 56.4& 90.4
\\
PruneVid{\small\texttt{(ACL2025)}}          & 53.0& \underline{55.7}& \underline{61.0}& 58.0& \underline{69.3}& 55.4& \underline{49.3}& \underline{56.9}& \underline{91.2}\\
FrameFusion{\small\texttt{(ICCV2025)}}  & 52.8& 53.0& 58.0& 56.7& 66.3& 55.4& 48.2& 55.1& 88.3
\\
MMG-Vid                                              & \textbf{54.9}& \textbf{56.3}& \textbf{63.4}& \textbf{59.4}& \textbf{71.0}& \textbf{57.9}& 49.2& \textbf{58.5}& \textbf{93.8}\\ \hline
\end{tabular}
}
\caption{\textbf{Comparison of state-of-the-art methods across video understanding benchmarks on LLaVA-Video-7B.} \textbf{Best} results are highlighted in bold, \underline{second best} underlined.}
\vspace{-3mm}
\label{tab:table1}
\end{table*}

\begin{table}[t]
\centering
\renewcommand{\arraystretch}{1.05} 
\begin{tabular}{l c c c c c} 
\hline
\multirow{2}{*}{Retention Ratio} & \multicolumn{2}{c}{Time (ms)}  & \multirow{2}{*}{Acc\%} 
\\ 
\cline{2-3}
& \multicolumn{1}{c}{Prefill} & \multicolumn{1}{c}{Generate} & \\ 
\hline 

LLaVA-OV     & 207.5 (1.0x)        & 329.7 (1.0x)         & 100 \\ 

25\%       & 52.8 (3.9x) & 107.2 (3.1x)     & 99.5 & \\ 
15\%   & 34.6  (6.0x) & 78.3 (4.2x)   & 98.3 & \\ 
\hline 
LLaVA-Video    & 277.5  (1.0x) & 411.6  (1.0x)   & 100 & \\ 
25\% & 65.0  (4.3x) & 142.9  (2.9x)   & 98.1 & \\ 

15\% & 42.5  (6.5x) & 109.6  (3.8x)   & 96.0 & \\ 
\hline
\end{tabular}
\caption{\textbf{Efficiency comparison of different Retention Ratios.} ``Prefill Time'': Time for model to generate first token; ``Generate Time'': Time for model to generate response.}
\vspace{-3mm}
\label{tab:table3}
\end{table}

\begin{table*}[pt]
\centering
\renewcommand{\arraystretch}{1.05}
\setlength{\tabcolsep}{1mm}
{
\begin{tabular}{c|ccccccc|cc}
\hline
\multirow{2}{*}{Method} & \multirow{2}{*}{MVBench} & \multirow{2}{*}{\makecell{LongVideo\\Bench}} & \multirow{2}{*}{MLVU} & \multicolumn{4}{c}{VideoMME} & \multicolumn{2}{|c}{\multirow{2}{*}{Average}} \\
\cline{5-8}
& & & & Overall & Short & Medium & Long & & \\
Duration & 16s & 1$\sim$60min & 3$\sim$120min & 1$\sim$60min & 1$\sim$3min & 3$\sim$30min & 30$\sim$60min & Score & \% \\ \hline
LLaVA-OneVision                      & 57.6& 56.6& 63.1& 58.5& 70.1& 56.6& 48.8& 59.0& 100.0\\  \hline
\multicolumn{10}{c}{\textit{\textbf{Retention Ratio: ~25\%}}} \\ \hline

FastV{\small\texttt{(ECCV24)}}              & 54.8& \textbf{56.8}& 59.3& 55.9& 66.0& 54.6& 47.2& 56.7& 96.1
\\
VisionZip{\small\texttt{(CVPR2025)}}        & \textbf{56.9}& 56.0& \underline{62.9}& \underline{58.0}& \underline{68.9}& \textbf{57.4}& 47.6& \underline{58.5}& \underline{99.2}\\
PruneVid{\small\texttt{(ACL2025)}}          & 55.7& 55.1& \textbf{63.4}& 57.0& 68.8& 54.4& 47.7& 57.8& 98.0
\\
FrameFusion{\small\texttt{(ICCV2025)}}  & 56.0& 54.8& 61.7& 57.5& 68.2& 55.7& \textbf{48.6}& 57.5& 97.5
\\
MMG-Vid                                               & \underline{56.7}& \underline{56.6}& \underline{62.9}& \textbf{58.6}& \textbf{71.2}& \underline{56.6}& \underline{48.1}& \textbf{58.7}& \textbf{99.5}\\ \hline
\multicolumn{10}{c}{\textit{\textbf{Retention Ratio: ~15\%}}} \\ \hline

FastV{\small\texttt{(ECCV24)}}                   & 52.3& 51.5& 56.3& 51.9& 58.4& 51.7& 45.4& 53.0& 89.8
\\
VisionZip{\small\texttt{(CVPR2025)}}        & \underline{55.7}& 54.2& 60.0& 55.5& 63.8& \underline{54.4}& \textbf{48.3}& 56.4& 95.6
\\
PruneVid{\small\texttt{(ACL2025)}}          & 55.0& \underline{55.6}& \textbf{61.9}& \underline{56.8}& \underline{67.9}& 54.3& 48.1& \underline{57.3}& \underline{97.1}
\\
FrameFusion{\small\texttt{(ICCV2025)}} & 55.1& 53.0& 58.3& 55.5& 65.8& 54.1& 46.7& 55.5& 94.1
\\
MMG-Vid                                               & \textbf{56.5}& \textbf{55.9}& \underline{61.6}& \textbf{57.9}& \textbf{69.6}& \textbf{56.0}& \underline{48.2}& \textbf{58.0}& \textbf{98.3}\\ \hline
\end{tabular}
}
\caption{\textbf{Comparison of state-of-the-art methods across video understanding benchmarks on LLaVA-OneVision-7B.}}
\vspace{-3mm}
\label{tab:table2}
\end{table*}

\section{Experiments}
\label{sec:experiments}
\subsection{Experimental Settings}

\noindent \textbf{Benchmarks and Baselines:} We employ lmms-eval~\cite{zhang2024lmms-eval} to evaluate our method on four widely-used video understanding benchmarks: MVBench~\cite{li2024mvbench}, LongVideoBench~\cite{wu2024longvideobench}, MLVU~\cite{zhou2024mlvu}, and VideoMME~\cite{fu2024videomme}. 
For comparison, we compare our method against several representative open-source methods, including FastV~\cite{chen2024fastv}, VisionZip~\cite{yang2025visionzip}, PruneVid~\cite{huang2024prunevid}, and FrameFusion~\cite{fu2024framefusion}.

\noindent \textbf{Implementation Details:} MMG-Vid is implemented into two VLLMs: LLaVA-OneVision-7B~\cite{li2024llava-ov} and LLaVA-Video-7B~\cite{zhang2024llava-video}. All experiments are conducted on NVIDIA H100-SXM5-80GB GPUs. Consistent with the official settings, we set the number of input frames to 32 for LLaVA-OneVision and 64 for LLaVA-Video. Furthermore, we set the similarity threshold $\tau$ to 0.95 and $\lambda$ to 0.5 for all experiments. For FastV, we prune tokens at the 2nd layer. For all intra-LLM methods, to ensure a fair comparison, we use the \textit{equivalent retention ratio} for token number calculation, which is defined as the average percentage of tokens processed across all layers of the LLM. For PruneVid, which compresses both visual tokens and the KV Cache, we evaluate only on its token compression strategy.

\subsection{Main Results}
To comprehensively evaluate the performance of our proposed MMG-Vid method, we conducted an extensive comparison against four advanced token compression methods on the LLaVA-Video and LLaVA-OneVision models. 

\noindent \textbf{(i) State-of-the-art Performance: } As detailed in Table \ref{tab:table1}, we established four distinct token Retention Ratios—{25\%, 20\%, 15\%, 10\%}—to thoroughly assess the performance of MMG-Vid and evaluate its robustness under varying compression intensity. The experimental results demonstrate that MMG-Vid consistently achieves optimal performance across all configurations, significantly outperforming all other baseline methods. Specifically, with 75\% of tokens compressed, MMG-Vid achieves a total average accuracy of \textbf{98.1\%}, whereas the best-performing baseline reached 96.8\%. As the compression ratio increases, the performance degradation of our method is substantially less pronounced than that of other methods, which demonstrates the efficacy of our marginal gain-based segment budgeting algorithm in dynamically adjusting the token budget to preserve the maximum visual information, especially under extreme compression intensity. When only 10\% of tokens are retained, our method still maintains an accuracy of \textbf{93.8\%}, a significant \textbf{16.7\%} higher than the 77.1\% of FastV, and outperforms the best baseline by \textbf{2.6\%}. Notably, the superiority of MMG-Vid is particularly evident on short videos across all compression levels (e.g., on MVBench, MMG-Vid is \textbf{2.53\%} and \textbf{1.86\%} higher than the best baseline at 20\% and 10\% retention ratio, respectively). We attribute this to the stronger temporal correlation between sampled frames in shorter videos. Our proposed MMG-Vid is designed to effectively leverage this temporal information, enabling it to achieve more comprehensive coverage of visual content and thus demonstrate superior performance. Besides, to assess the cross-model robustness of MMG-Vid, we extended our evaluation to the LLaVA-OneVision, as detailed in Table~\ref{tab:table2}. The results reveal that MMG-Vid still maintains a consistent superiority over all competing methods at various compression rates. Impressively, it exhibits negligible performance degradation at retention ratios of 25\% and 15\%, preserving 99.5\% and 98.3\% of the baseline performance, respectively. 

\noindent \textbf{(ii) Superior Inference Efficiency:} Beyond Performace, Table \ref{tab:table3} presents comprehensive real-world inference latency. MMG-Vid achieves a \textbf{3.9x }acceleration in the prefill phase and a \textbf{3.1x} acceleration in the generation phase, while maintaining \textbf{99.5\% }of the model performance on LLaVA-OneVision,  significantly reducing computational costs.

\begin{table}[t]
\centering
\renewcommand{\arraystretch}{1.05} 
\setlength{\tabcolsep}{1mm}
\begin{tabular}{c|cccccc}
\hline
\multirow{2}{*}{\makecell{$\lambda$ in \\ Budgeting}} & \multirow{2}{*}{MLVU} & \multicolumn{4}{c}{VideoMME} & \multirow{2}{*}{Acc \% } \\
\cline{3-6}
 & & Overall & Short & Medium & Long & \\
\hline
Vanilla & 67.4 & 64.3 & 77.3 & 62.3 & 53.2 & 100.0 \\ 
\hline
1.0 & 65.2 & 62.0 & 74.3 & 60.2 & 51.3 & 96.6
\\
0.8 & 65.5 & 61.9& 74.2& 60.2 & 51.1 & 96.7
\\
0.6 & \textbf{66.2}& 62.0& 74.3 & 59.7 & 51.9 & 97.3
\\
0.5 & 66.2& \textbf{62.3}& \textbf{74.7} & 60.0 & \textbf{52.1}& \textbf{97.6}
\\
0.4 & 65.9& 61.8& 74.3 & 59.8 & 51.3 & 97.0\\
0.2 & 65.4& 61.6& 74.1 & 59.6 & 51.0 & 96.4
\\
0 & 66.0& 62.0& 74.2& \textbf{60.3}& 51.3& 97.2
\\
\hline
\end{tabular}
\caption{ \textbf{Ablation study on $\lambda$ for marginal gain-based segment budgeting on LLaVA-Video (Retention Ratio: 25\%).} Our method consists of Representiveness and Diversity, $\lambda$ determines the proportion between them.}
\vspace{-3mm}
\label{tab:table4}
\end{table}

\begin{figure}[t]
    \centering
    \includegraphics[width=\linewidth]{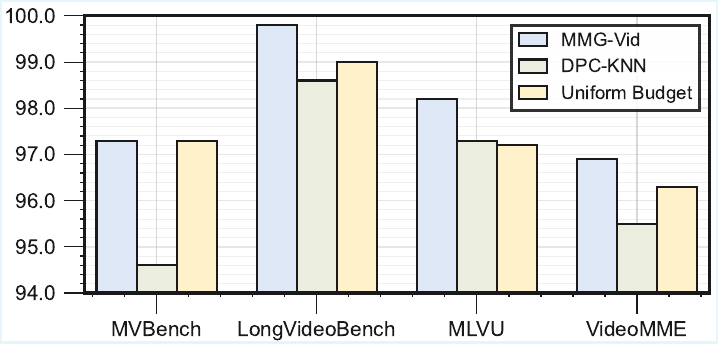}
    \vspace{-2mm}
    \caption{\textbf{Ablation study of MMG-Vid's modules on LLaVA-Video (Retention Ratio: 25\%).} ``DPC-KNN'' refers to using the standard DPC-KNN algorithm instead of our proposed TG-DPC. ``Uniform Budget'' refers to the conventional method of assigning a fixed budget to each frame. }
    \vspace{-3mm}
\label{fig:ablation_study}
\end{figure}

\subsection{Ablation study}
\subsubsection{Ablation study on different modules}

To validate the effectiveness of our two strategies, we perform ablation experiments on the LLaVA-Video model with a 25\% retention ratio. The results are presented in Figure \ref{fig:ablation_study}. Replacing our TG-DPC with a standard DPC-KNN algorithm leads to a significant performance degradation across all applicable benchmarks. This is particularly evident on short-video benchmarks like MVBench, where performance drops from 97.3\% to 94.6\%, underscoring the superiority of our TG-DPC in capturing complex temporal dynamics compared to conventional methods. Furthermore, compared to a uniform static budget allocation, our dynamic approach boosts performance by 0.8\% and 1.0\% on long-video benchmarks such as LongVideoBench and MLVU, respectively. This demonstrates that our method can judiciously allocate the budget to maximize information density coverage in long videos, which typically feature more diverse segments.

\subsubsection{Ablation study on budgeting}
Our marginal gain-based segment budgeting module balances two components: Representativeness (selecting core content) and Diversity (reducing redundancy), controlled by the hyperparameter $\lambda$. As shown in Table \ref{tab:table4}, the Diversity-only setting outperforms the Representativeness-only setting by 0.6\%. Nevertheless, the optimal performance is achieved through a balanced combination. The peak scores on MLVU and VideoMME are obtained with a set to 0.6 and 0.5, respectively. This demonstrates that a balanced integration of both modules is crucial for preserving the most comprehensive video information.

\section{Conclusion}
In this paper, we introduce MMG-Vid, a novel training-free video token pruning framework that significantly reduced the computational cost of various Video Large Language Models (VLLMs). Unlike prior methods that relied on static uniform budgeting and disjoint pruning, our MMG-Vid formulates the token pruning task as a constrained subset selection problem and optimizes the selected token subset by consistently maximizing marginal information gain. We achieve more comprehensive coverage of visual representation by first segmenting videos into semantically coherent segments, then dynamically allocating budgets based on contextual relevance, and finally performing unified spatiotemporal pruning via our proposed TG-DPC algorithm. Extensive experimental results across multiple VLLMs and benchmarks demonstrate that MMG-Vid effectively reduce the inference latency while maintaining superior performance, enabling the practical deployment of VLLMs.

\clearpage
\bibliography{aaai2026}

\clearpage
\appendix

\section*{Appendix}
In the appendix, we provide more benchmark, model and baseline details in Experiments.

\section{Benchmark Details}
\label{sec:appendix/benchmarks}
We evaluate MMG-Vid on various multi-modal understanding benchmarks detailed as follows:
\begin{itemize}
 \item \textbf{MVBench}~\cite{li2024mvbench} formulates 20 video understanding tasks, each with 200 QA pairs, to assess temporal comprehension beyond single-frame analysis and provide a comprehensive model evaluation.
  \item \textbf{LongVideoBench}~\cite{wu2024longvideobench} consists of 3,763 videos and 6,678 associated multiple-choice questions,  covering diverse domains such as movies, news, specifically designed to assess a model's capacity for temporal information retrieval and analysis.
 \item \textbf{MLVU}~\cite{zhou2024mlvu} designed for long-form videos, features a benchmark with durations from 3 minutes to over 2 hours (12-minute average). It covers diverse genres like movies, documentaries, TV series, while evaluating models across 9 distinct tasks such as topic reasoning, video summarization, and needle question answering.
 \item \textbf{VideoMME}~\cite{fu2024videomme} consists of 900 videos and 2,700 QA pairs with durations from 11 seconds to 1 hour.These are categorized into three temporal subsets (short-, medium-, and long-term) and span 6 main visual domains, such as life record and knowledge.
\end{itemize}

\section{Model Details}
\label{sec:appendix/models}
The performance of our model is evaluated against two baseline VLLMs, which are detailed below: 
\begin{itemize}
    \item \textbf{LLaVA-OneVision}~\cite{li2024llava-ov} unifies single-image, multi-image, and video tasks within a single model by representing video as a long sequence of visual tokens. This representation facilitates a seamless transfer from image to video tasks and enhances its robust zero-shot video understanding capabilities. Furthermore, bilinear interpolation is employed to reduce the number of tokens, allowing the consideration of a larger number of frames by reducing tokens per frame. 
    \item \textbf{LLaVA-Video}~\cite{zhang2024llava-video} builds upon the single-image stage checkpoint of LLaVA-OneVision and is subsequently fine-tuned on a large synthetic video instruction dataset (LLaVA-Video-178K). Architecturally, the model employs the SigLIP visual encoder and Qwen2 as its Large Language Model, introducing newline tokens for each frame to distinguish spatial and temporal positions and thereby achieving robust video comprehension across various benchmarks.

\end{itemize}

\section{Baseline Details}
\label{sec:appendix/baselines}
A detailed analysis and comparison of the token compression methods mentioned in this paper is presented as follows: 
\begin{itemize}
    \item \textbf{FastV}~\cite{chen2024fastv} first reveals the high redundancy of visual tokens in language models, and performs token pruning at the K-th layer of the LLM based on attention scores, effectively reducing computational overhead intra-LLM. However, its explicit dependence on attention weights makes it incompatible with efficient attention. In our experimental setup, we re-implement the method and consistently set K to 2 across all conducted experiments.

    \item \textbf{VisionZip}~\cite{yang2025visionzip} selects important tokens using the average attention each token receives from all others in the sequence and then merges the remaining ones. However, this method conflicts with pooling operations in Video LLMs, leading to performance degradation. We address this by applying pooling to the attention weights and key-values before the pruning stage. Following the original settings, each frame preserves dominant and contextual tokens in a 54:10 ratio.
    \item \textbf{PruneVid}~\cite{huang2024prunevid} clusters frames into segments and classifies tokens within each segment as either static or dynamic, applying different pruning strategies accordingly. While the original PruneVid compresses both visual tokens and KV Cache, to ensure a fair comparison, we only implement the pre-LLM pruning. Following the original configuration, we set the threshold $\tau$ =  0.8 and the temporal segment ratio $\gamma$ = 0.25.
    \item \textbf{FrameFusion}~\cite{fu2024framefusion} is motivated by two key observations. First, spatially corresponding visual tokens between adjacent frames exhibit significantly higher cosine similarities than other token pairs. Second, although  the similarity between highly similar tokens decreases at deeper layers, the relative similarity rankings of these tokens remain stable. Inspired by this strong consistency, FrameFusion adopts a two-stage approach: it first merges tokens across frames and subsequently selects important ones. Specifically, it employs a similarity-based merging strategy that operates in a cascaded manner; once highly similar tokens are merged at shallower layers, they remain unified throughout subsequent computations. Following this merging phase, a final pruning step retains only the most important tokens, selected according to their attention scores. Following the original configuration, we set the $S_{\text{threshold}}$ = 0.6 and the $N_{\text{threshold}}$ = 0.1. 
\end{itemize}
\end{document}